\documentclass[letterpaper]{article} 
\usepackage{aaai24}  
\usepackage{times}  
\usepackage{helvet}  
\usepackage{courier}  
\usepackage[hyphens]{url}  
\usepackage{graphicx} 
\usepackage{natbib}  
\usepackage{caption} 
\usepackage{algorithm}
\usepackage{algorithmic}
\usepackage{amsmath}
\usepackage{amssymb}
\usepackage{booktabs}

\usepackage{newfloat}
\usepackage{listings}
\usepackage{xcolor}
\DeclareCaptionStyle{ruled}{labelfont=normalfont,labelsep=colon,strut=off} 
\floatstyle{ruled}
\newfloat{listing}{tb}{lst}{}
\floatname{listing}{Listing}
\title{Quantum Sparse Autoencoders for Q-Matrix Estimation in Cognitive Diagnosis}
\author{
    Arif Hassan Zidan\textsuperscript{\rm 1},
    Yi Pan\textsuperscript{\rm 2},
    Bowen Guo\textsuperscript{\rm 3},
    Xiang Li\textsuperscript{\rm 5},
    Yu Bao\textsuperscript{\rm 3},
    Yingfeng Wang\textsuperscript{\rm 4}, \\
    Tianming Liu\textsuperscript{\rm 2},
    Wei Zhang\textsuperscript{\rm 1}\thanks{Corresponding author.}
}
\affiliations{
    \textsuperscript{\rm 1}School of Computer and Cyber Sciences, Augusta University, Augusta, GA, USA\\
    \textsuperscript{\rm 2}School of Computing, University of Georgia, Athens, GA, USA\\
    \textsuperscript{\rm 3}Department of Graduate Psychology, James Madison University, Harrisonburg, VA, USA\\
    \textsuperscript{\rm 4}Department of Computer Science and Engineering, University of Tennessee at Chattanooga, Chattanooga, TN, USA\\
    \textsuperscript{\rm 5}Department of Radiology, Massachusetts General Hospital, Harvard Medical School, Boston, MA, USA\\
}

\begin{document}

\maketitle

\begin{abstract}
Q-matrices play a central role in cognitive diagnosis within educational data mining (EDM), specifying which latent skills each assessment item requires. Data-driven Q-matrix estimation remains challenging when assessments involve many correlated skills and when real response patterns depart from idealized generative assumptions. We introduce a novel \emph{quantum sparse autoencoder} (QSAE) for Q-matrix estimation, which, to the best of our knowledge, is the first application of quantum machine learning (QML) to cognitive diagnosis. Overall, the QSAE embeds each student's binary response vector into a quantum circuit using an encoder, compresses it into a sparse latent representation, and maps that representation to the Q-matrix. We benchmark the QSAE against a classical autoencoder (CAE) across 60 simulated datasets and 9 real-world assessment datasets. The results reveal complementary strengths. Although the CAE partially achieves higher average accuracy under several simulation conditions, the QSAE is substantially more stable across replications, exhibiting lower variance in 49 of the 60 conditions.  Moreover, on real assessment data, the QSAE outperforms the CAE on 6 of the 9 datasets. These findings suggest that the principal advancement of QML in this setting is not universal accuracy improvement, but enhanced robustness and capability to explore latent-structure complexity in real datasets. 
\end{abstract}

\section{Introduction}
\label{sec:intro}
Quantum machine learning (QML) has emerged as an active research frontier at the intersection of quantum computing and artificial intelligence, motivated by the possibility that quantum systems can represent and process information in ways that are difficult to reproduce classically~\cite{cerezo2022challenges, jiang2026qaisurvey}. Encoding classical data into an $n$-qubit quantum state provides access to a $2^n$-dimensional Hilbert space, creating an exponentially large representation space in which complex structures that are difficult to separate in the original feature space may become more amenable to learning~\cite{havlicek2019supervised}. On current noisy intermediate-scale quantum (NISQ) devices, much of this potential is explored through hybrid quantum--classical variational algorithms~\cite{cerezo2021vqa, pan2025molqae, jahin2025quantum, pan2026symmetry}. 

QML is anticipated to be particularly valuable in domains characterized by complex, high-dimensional, and structured observations, such as drug discovery, materials science, and scientific data analysis~\cite{jiang2026qaisurvey}. A recurring challenge in these domains is \emph{representation learning}: transforming high-dimensional observations into compact latent representations that preserve the information most relevant to downstream inference. Quantum autoencoders (QAEs) provide a natural mechanism for this purpose by using parameterized quantum circuits to compress input states into lower-dimensional latent subsystems~\cite{pan2025molqae}. Importantly, this compression principle is not inherently domain-specific. Whenever observed data are governed by a relatively small number of hidden factors, learning a compact latent representation may provide a useful route to recovering the underlying structure. Educational assessment provides a particularly compelling example of such a setting. Therefore, we adopt this paradigm to investigate whether quantum representation learning can reveal latent structure in educational assessment data.

Traditionally, in educational measurement, diagnostic classification models (DCMs)~\cite{henson2009defining, rupp2010diagnostic, bradshaw2016diagnostic, lin2025comprehensive} infer students' mastery of latent skills, or \emph{attributes}, from their observed item responses. A fundamental component of a DCM is the \emph{Q-matrix} ~\cite{tatsuoka1983rule}, a binary matrix in which entry $(j,k)$ indicates whether item $j$ requires attribute $k$. The Q-matrix therefore defines the item--attribute dependency structure on which subsequent diagnostic inference depends. In practice, however, Q-matrices are commonly specified manually by domain experts, making the process costly, labor-intensive, and potentially subjective ~\cite{pongsophon2026beyond}. For example, on a widely used Trends in International Mathematics and Science Study (TIMSS) mathematics assessment, two expert coders agreed on only approximately $89\%$ of Q-matrix entries~\cite{li2021rbm}. These limitations have motivated increasing interest in estimating Q-matrices accurately.

Data-driven Q-matrix estimation nevertheless remains challenging. The search space grows exponentially with the number of items and attributes; the latent attributes are unobserved and identifiable only up to permutation; educational datasets are often modest in size; and correlations among attributes can obscure the true item--attribute relationships~\cite{li2021rbm, islam2025integration}. Conventional approaches can perform well when the assumed generative structure closely matches the data, but their performance may deteriorate when real-world responses deviate from idealized model assumptions. These challenges motivate the exploration of alternative representation learning with different inductive biases.

In this work, we develop an innovative quantum sparse autoencoder (QSAE) to provide an effective alternative for data-driven Q-matrix estimation. We introduce a QSAE framework that first embeds each student's binary response vector into an eight-qubit quantum circuit using a response-to-quantum-state encoding specifically designed for educational assessment data. A parameterized quantum encoder then compresses the response information into a $K$-dimensional sparse latent representation intended to capture the underlying skill structure, from which the Q-matrix is subsequently recovered. The imposed sparsity encourages the QSAE to retain the most informative latent features while suppressing noise and spurious variation, potentially improving robustness when applied to real-world response data. The encoder--decoder architecture extends the MolQAE framework~\cite{pan2025molqae} from molecular representation learning to educational assessment data. To the best of our knowledge, this is the first study to apply a quantum autoencoder to Q-matrix estimation and the first use of quantum representation learning for cognitive diagnosis.

Rather than assuming that the quantum model should universally outperform its classical counterpart, we ask a more informative question: \emph{under what data conditions does each representation-learning approach provide an advantage?} To answer this question, we benchmark the proposed QSAE against a classical autoencoder (CAE) following the evaluation protocol of~\cite{li2021rbm, ramospulido2025qmledu}. The comparison covers $60$ simulated conditions generated under the Deterministic Inputs, Noisy ``And'' gate (DINA) model~\cite{najera2023restricted, gu2019identifiability} and $9$ real-world assessment datasets. Performance is evaluated using standard Q-matrix recovery criteria, including overall error (OE), out-of-true-positive error (OTP), and out-of-true-negative error (OTN)~\cite{li2021rbm}.

The results reveal a complementary pattern. Under clean simulated conditions, where the data-generating structure is well aligned with the inductive bias of the classical model, the CAE achieves higher average recovery accuracy, consistent with expectations from prior representative work~\cite{li2021rbm}. However, the QSAE exhibits substantially greater stability across replications, producing lower standard deviations in $49$ of the $60$ simulated conditions. More importantly, across the $9$ real datasets, the QSAE outperforms the CAE on $6$ datasets. These findings suggest that the primary advantage of the proposed QSAE is not universal superiority under idealized conditions, but rather its robustness to sampling variability and model misspecification in more realistic settings.

Overall, this work makes the following contributions:
\begin{itemize}
    \item \textbf{A quantum sparse autoencoder framework for Q-matrix estimation.}
    We introduce, to the best of our knowledge, the first QSAE-based approach to Q-matrix recovery and the first application of quantum representation learning to cognitive diagnosis. 

    \item \textbf{A systematic quantum--classical benchmark.}
    We conduct a controlled comparison between the proposed QSAE and a classical autoencoder across $60$ simulated DINA conditions and $9$ real assessment datasets using standard Q-matrix recovery metrics (OE, OTP, and OTN)~\cite{li2021rbm}.

    \item \textbf{An empirical characterization of complementary strengths.}
    We show that the CAE achieves stronger recovery accuracy under well-specified simulations, whereas the QSAE provides greater replication stability and stronger performance across real datasets. 
\end{itemize}

\section{Related Work}
\label{sec:related}

\paragraph{Q-matrix estimation.}
Because manually specifying a Q-matrix is costly, labor-intensive, and potentially error-prone, a substantial body of research has focused on estimating or validating Q-matrices directly from response data. Early data-driven approaches include likelihood-based estimation~\cite{wang2020q} and Bayesian methods that place prior distributions over the Q-matrix and estimate its structure using Markov Chain Monte Carlo (MCMC) sampling~\cite{chung2018mcmc, haertel1989using}. 

More recently, representation-learning approaches have been introduced for Q-matrix recovery. Investigators~\cite{li2021rbm} utilized a Restricted Boltzmann Machine (RBM) to estimate large Q-matrices and showed that, under the DINA model, a main-effects formulation can recover the required skill structure. Related studies have explored sparse and constraint-based autoencoders for identifying item--skill relationships in item response theory and cognitive diagnosis settings~\cite{paassen2022sparse, ramospulido2025qmledu}. In this work, we adopt the benchmark protocol and the OE, OTP, and OTN recovery metrics introduced by~\cite{li2021rbm}, while using their RBM results as an additional reference point for comparison.

Despite these advances, existing approaches remain entirely classical, and their performance is generally strongest when the assumed model structure is well aligned with the underlying data-generating process. We therefore introduce a quantum representation-learning counterpart and systematically investigate how quantum and classical approaches behave both when this structural assumption is satisfied and when real-world data depart from it.

\paragraph{Quantum representation learning.}
Quantum autoencoders (QAEs) were introduced for the efficient compression of quantum data, encoding an input state into a smaller latent register while disentangling the complementary ``trash'' qubits into a fixed reference state~\cite{romero2017qae}. Subsequent work has extended QAEs toward classical domain data: MolQAE encodes complete molecular structures directly from their sequential (SMILES) descriptions and reconstructs them with high fidelity under substantial dimensionality reduction~\cite{pan2025molqae}. Our pipeline extends this framework from molecular tokens to student response vectors, adapting its state-preparation and compression method for educational assessment. Additionally, QML has recently reached educational data, and prior work has targeted only predictive data-mining tasks rather than psychometric measurement, for example classifying alumni career outcomes with quantum-kernel support vector machines~\cite{ramospulido2025qmledu}. To our knowledge, no prior work applies quantum models to Q-matrix estimation or, more broadly, to the latent-structure measurement models at the core of cognitive diagnosis.

%
\section{Preliminaries}
\label{sec:prelim}

Q-matrix estimation can be viewed as an unsupervised latent-structure
recovery problem, closely related to feature extraction and matrix
factorization~\cite{chuong2025decomposition}. We first introduce the two central objects in cognitive
diagnosis: the \emph{response matrix}, which contains the observed data,
and the \emph{Q-matrix}, which represents the latent item--skill structure
to be recovered.

\paragraph{Response matrix.}
The response matrix is the input data. Consider an assessment administered to $N$ students, each responding to
$J$ items. The observed data are represented by a binary
\emph{response matrix}, $R \in \{0,1\}^{N \times J}$, where $R_{ij}=1$ if student $i$ answers item $j$ correctly and $R_{ij}=0$ otherwise. Each row of $R$ therefore represents a student's response pattern across all assessment items. The response matrix is directly observed, whereas the underlying skills that give rise to these
responses remain latent.

\paragraph{Q-matrix.}
The Q-matrix represents the weight matrix. Each assessment item is assumed to require a subset of $K$ latent
\emph{skills}, also referred to as \emph{attributes}. The
\emph{Q-matrix}, $Q \in \{0,1\}^{J \times K}$, encodes this item--skill dependency structure, where $q_{jk}=1$ indicates that item $j$ requires skill $k$, and $q_{jk}=0$ otherwise. Thus, the $j$-th row of $Q$ specifies the set of skills required by item $j$.

From a latent-factor perspective, the Q-matrix plays a role analogous to
a sparse loading or mixing matrix that links observed items to underlying
latent factors. Conventional educational assessment uses $Q$ with
the observed responses $R$ to infer each student's skill-mastery profile.
Consequently, accurate specification of $Q$ is essential for valid
diagnostic inference. 

\paragraph{Data-generating model.}
For the simulation experiments, response data are generated using the
DINA (Deterministic Inputs, Noisy ``And'' gate) model, a widely used
cognitive diagnosis model~\cite{junker2001cognitive,li2021rbm}.
Under DINA, a student is expected to answer an item correctly only if the
student has mastered \emph{all} skills required by that item, subject to
item-specific guessing and slipping parameters. The DINA model therefore
induces a well-defined conjunctive relationship between latent skills and
observed responses. Importantly, the DINA model is used only to generate controlled
simulation data and provide a known ground-truth Q-matrix for evaluation.
Neither the proposed QSAE nor the classical baseline is provided with the
true $Q$ or explicitly informed of the underlying data-generating model.

\paragraph{Task formulation.}
Given only the observed response matrix $R$, our goal is to recover an
estimate $\hat{Q} \in \{0,1\}^{J \times K}$ of the underlying Q-matrix. This formulation can be interpreted as an unsupervised latent-structure learning problem in which the item--skill relationships must be inferred from response patterns alone.


\section{Methodology}
\label{sec:method}
\begin{figure*}[t]
  \centering
  \includegraphics[width=0.75\textwidth]{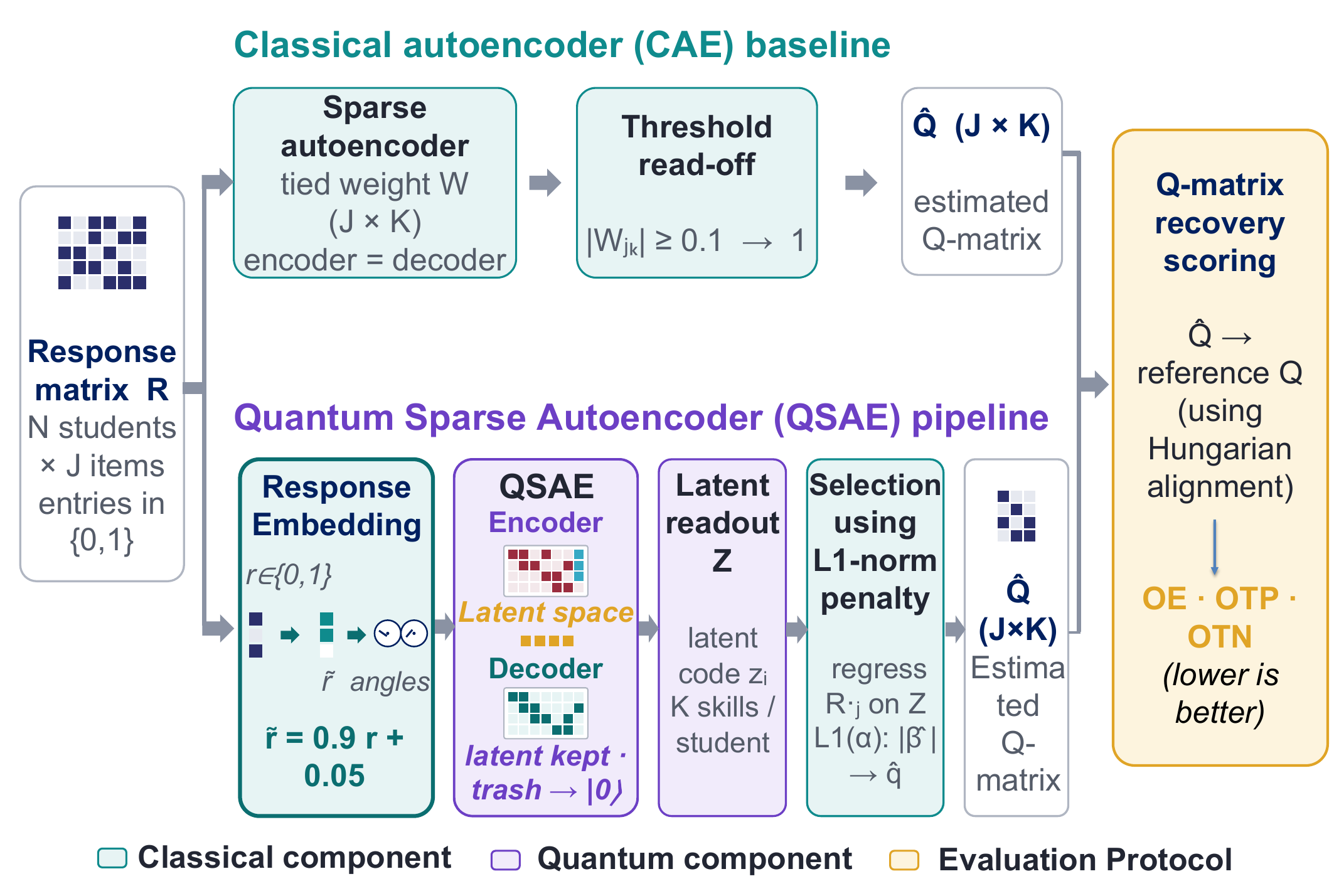}
  \caption{\textbf{Overview of the Q-matrix estimation pipeline.} Both
    estimators take the same binary response matrix $R \in \{0,1\}^{N
    \times J}$ as input and produce an estimated Q-matrix
    $\hat{Q} \in \{0,1\}^{J \times K}$, scored against the reference $Q$
    under a common protocol. \emph{Top (classical baseline, CAE):} a
    sparse tied-weight autoencoder is trained on $R$, and its weight
    matrix is thresholded ($|W_{jk}| \ge 0.1$) into a binary $\hat{Q}$.
    \emph{Bottom (quantum pipeline, QSAE):} responses are embedded as
    rotation angles ($\tilde{r} = 0.9\,r + 0.05$), compressed by a quantum
    autoencoder whose trash qubits are driven to $\lvert 0 \rangle$, read
    out as a $K$-dimensional latent code $z_i$ per student, and turned into
    $\hat{Q}$ by per-item $L_1$-penalized regression of each response
    column on the latent readout. \emph{Right (evaluation):} both estimates
    are aligned to $Q$ by the Hungarian algorithm and scored with OE, OTP,
    and OTN (lower is better). Teal blocks are classical components,
    purple are quantum, and the yellow block is the evaluation protocol.}
  \label{fig:pipeline}
\end{figure*}

Figure~\ref{fig:pipeline} gives an overview of the two estimators, CAE and QSAE. Both take the binary response matrix as input and produce a Q-matrix estimate scored against the reference; they differ only in how the latent skill structure is learned. In this section, we describe the shared problem setup and then each pipeline in turn.

\subsection{Embedding Responses as Quantum States}
\label{sec:method:encoding}
 To process responses on a quantum circuit, each student's binary response
vector must be mapped to the parameters of a quantum state. Following the
common angle-encoding strategy, in which classical features set the
rotation angles of parameterized single-qubit
gates~\cite{larose2020robust}, we encode a
response vector $r_i \in \{0,1\}^{J}$ into single-qubit rotation angles.
Binary values sit at the extremes of the rotation range, where $0$ and
$\pi$ produce degenerate (indistinguishable) states and vanishing
gradients; feeding raw $0/1$ values directly is therefore numerically
poor, an instance of the broader observation that the choice of data
encoding materially affects a quantum model's
trainability~\cite{schuld2021encoding}. We instead apply
an affine smoothing,
\begin{equation}
  \tilde{r}_{ij} = 0.9\, r_{ij} + 0.05,
  \label{eq:smoothing}
\end{equation}
which maps $0 \mapsto 0.05$ and $1 \mapsto 0.95$, keeping every angle
strictly inside the usable range and away from the degenerate endpoints
while preserving the binary contrast. The smoothed values parameterize the
rotation gates that prepare the input state. This response-to-angle embedding is the component we design specifically for educational assessment data; it is the educational-domain counterpart of the molecular state preparation used by MolQAE~\cite{pan2025molqae}.

\subsection{QSAE Architecture}
\label{sec:method:qae}
 
Our QSAE is the quantum analogue of a classical
autoencoder, which learns a compressed latent representation by training a
network to reconstruct its input~\cite{hinton2006reducing}. It follows
the encoder-compression-decoder design of
MolQAE~\cite{pan2025molqae}, and is trained as a parameterized quantum circuit, i.e., a machine-learning model~\cite{benedetti2019parameterized}. The circuit operates on
$n$ qubits, partitioned into a \emph{latent} register that retains the
compressed representation and a set of \emph{trash} qubits that are driven
toward a fixed reference state.
 
\paragraph{Encoder.}
The smoothed responses from Eq.~\eqref{eq:smoothing} set the rotation
angles of a parameterized input layer, preparing a data-dependent input
state. A trained encoder circuit $U_{\text{enc}}(\theta)$, built from
parameterized single-qubit rotations and entangling gates, transforms this
state so that the information needed to reconstruct the input is
concentrated in the latent register.
 
\paragraph{Compression via trash qubits.}
Compression is achieved by pushing the trash qubits toward a fixed
reference state $\lvert 0 \rangle$. When the trash qubits are successfully
disentangled and reset to $\lvert 0 \rangle$, all reconstruction-relevant
information must reside in the latent register, which therefore forms a
compact latent code of the response
pattern~\cite{pan2025molqae}.

\paragraph{Sparse representation.}
The $L_1$ norm, $\lVert \beta \rVert_1 = \sum_k |\beta_k|$, is the
standard tool for inducing sparse representations in linear
models~\cite{hastie2015sparsity}. Added to a
least-squares objective it yields the
LASSO~\cite{tibshirani1996lasso}, which shrinks coefficients and drives
the uninformative ones exactly to zero, so the fitted model retains only a
small subset of active predictors; efficient solvers make this practical
even for high-dimensional, repeatedly solved
problems~\cite{liu2009slep}. Unlike the $L_2$ (ridge) penalty, which
shrinks all coefficients smoothly but leaves them non-zero, the $L_1$
penalty performs genuine variable selection: it identifies \emph{which}
predictors matter rather than merely how much to down-weight
them~\cite{hastie2015sparsity}. This suits Q-matrix recovery, where each
item loads on only a few of the $K$ skills, so the target coefficient
vector is sparse and selecting its non-zero entries directly identifies
the required skills.

\paragraph{Decoder.}
A decoder circuit $U_{\text{dec}}$ maps the latent register (with the
reset trash qubits) back toward the input state. The model is trained to
maximize reconstruction fidelity while enforcing the compression
constraint, giving the objective
\begin{equation}
  \mathcal{L}
    = \bigl(1 - F_{\text{recon}}\bigr)
      + \lambda_{\text{t}}\, D_{\text{trash}},
  \label{eq:qae-loss}
\end{equation}
where $F_{\text{recon}}$ is the reconstruction fidelity between the input
and reconstructed states, $D_{\text{trash}}$ penalizes deviation of the
trash qubits from $\lvert 0 \rangle$, and $\lambda_{\text{t}}$ balances the
two terms (we use $\lambda_{\text{t}} = 0.1$). Training on the response
matrix requires no knowledge of the Q-matrix: the QSAE is a purely
unsupervised model of the responses, and $Q$ enters only at evaluation.

\subsection{Q-matrix Recovery}
\label{sec:method:recovery}
 
The trained autoencoder yields, for each student $i$, a $K$-dimensional
latent readout $z_i \in \mathbb{R}^{K}$ obtained from the latent register.
Stacking these gives a latent matrix $Z \in \mathbb{R}^{N \times K}$ whose
columns act as data-driven surrogates for the $K$ latent skills. It
remains to decide, for each item, which skills it depends on: this is a
variable-selection problem.
 
\paragraph{Sparse per-item selection.}
The key structural fact is that Q-matrices are sparse: each item loads on
only a few skills. \cite{li2021rbm} exploit exactly this property, using
an $L_1$ penalty so that the sparse (non-zero) structure of a learned
weight matrix reveals the Q-matrix. We adopt the same principle at the
recovery stage. For each item $j$, we regress its response column
$R_{\cdot j}$ on the latent skills $Z$ with an $L_1$-penalized (LASSO)
model~\cite{liu2009slep, tibshirani1996lasso},
\begin{equation}
  \hat{\beta}_j
    = \arg\min_{\beta \in \mathbb{R}^{K}}
      \; \bigl\lVert R_{\cdot j} - Z\beta \bigr\rVert_2^2
      + \alpha \lVert \beta \rVert_1 ,
  \label{eq:lasso}
\end{equation}
where the $L_1$ term shrinks the coefficients of irrelevant skills toward
zero and retains only the skills that carry signal for item $j$. The
magnitude $\lvert \hat{\beta}_{jk} \rvert$ scores how strongly item $j$
depends on skill $k$. Recovering the full Q-matrix is thus a collection of
$J$ related sparse-regression problems over a shared latent skill
representation.
 
\paragraph{Binarization.}
The coefficient magnitudes are converted to binary Q-matrix entries by thresholding, as is standard in Q-matrix learning: \cite{li2021rbm}
recover their Q-matrix by thresholding the magnitudes of the learned
weights. We apply the same principle to the LASSO coefficients. A fixed sparsity level is a reasonable prior given that real Q-matrices are sparse and low-order, and it parallels the cutoff used by \cite{li2021rbm} and our CAE.

\subsection{CAE Baseline}
\label{sec:method:classical}
 
As a classical counterpart, we use a sparse autoencoder with a single
tied weight matrix $W \in \mathbb{R}^{J \times K}$ shared by the encoder
and decoder, so that $W_{jk}$ directly represents the strength of the
association between item $j$ and skill $k$. The model is trained to
reconstruct the response matrix under an $L_1$ penalty on $W$, which,
following the same sparsity principle as above~\cite{li2021rbm,%
tibshirani1996lasso}, drives uninformative item--skill weights toward
zero. The Q-matrix is then read off with the threshold of
\cite{li2021rbm}: entries with $\lvert W_{jk} \rvert \ge 0.1$ are set to
$1$ and the rest to $0$. Because its tied-weight, additive structure
matches the bilinear item--skill form assumed by additive CDMs, this
baseline is a strong and well-motivated point of comparison.

\subsection{Evaluation Protocol}
\label{sec:method:eval}
 
We follow the evaluation protocol of \cite{li2021rbm}.
 
\paragraph{Column alignment.}
Because the latent skills are recovered only up to a permutation, the
columns of $\hat{Q}$ do not necessarily correspond to those of the
reference $Q$ in order. We resolve this by matching columns with the
Hungarian algorithm~\cite{wang2020q}, which finds the column
permutation minimizing the total disagreement between $\hat{Q}$ and $Q$
before any error is computed.
 
\paragraph{Evaluation metrics.}
After alignment, we report three standard Q-recovery error rates, for all
of which \emph{lower is better}. The \emph{overall error} (OE) is the fraction of all $J \times K$ entries of $\hat{Q}$ that disagree with $Q$. Because Q-matrices are sparse, OE alone can look small even for a poor estimate that predicts mostly zeros, so we also report two complementary rates that expose the two failure modes separately. The \emph{out-of-true-positive} error (OTP) is the fraction of true $1$-entries that were missed
(predicted $0$), measuring under-specification (required skills not
detected). The \emph{out-of-true-negative} error (OTN) is the fraction of
true $0$-entries wrongly predicted as $1$, measuring over-specification
(spurious skills attached to items). OE, OTP, and OTN are defined as:
\begin{equation}
\mathrm{OE} = \frac{1}{JK}\sum_{j=1}^{J}\sum_{k=1}^{K}
\mathbb{1}\{\hat{q}_{jk} \neq q_{jk}\},
\end{equation}
\begin{equation}
\mathrm{OTP} = \frac{\sum_{j,k}\mathbb{1}\{\hat{q}_{jk}=0,\;
q_{jk}=1\}}{\sum_{j,k}\mathbb{1}\{q_{jk}=1\}},
\end{equation}
\begin{equation}
\mathrm{OTN} = \frac{\sum_{j,k}\mathbb{1}\{\hat{q}_{jk}=1,\;
q_{jk}=0\}}{\sum_{j,k}\mathbb{1}\{q_{jk}=0\}}.
\end{equation}

Reporting OE, OTP, and OTN together keeps both missed and spurious loadings visible, following standard practice in Q-matrix recovery and validation~\cite{delatorre2016validation,li2021rbm}.

\section{Results}
\label{sec:results}

\subsection{Experimental Setup}
\label{sec:results:setup}

We evaluate both autoencoders on simulated and real data, following the
benchmark protocol of~\cite{li2021rbm}. The simulated benchmark spans
$60$ conditions generated under the DINA model, crossing the number of
skills $K \in \{5, 10, 15, 20, 25\}$, sample size
$N \in \{2000, 10000\}$, skill correlation
$\rho \in \{0, 0.25, 0.75\}$, and guess/slip noise
$g = s \in \{0.1, 0.2\}$, with $J = 3K$ items and $5$ independent
replications per condition. The real benchmark comprises $9$ publicly
available assessment datasets with expert-constructed Q-matrices. All
estimates are aligned to the reference Q-matrix by the Hungarian algorithm
and scored with overall error (OE), out-of-true-positive error (OTP), and
out-of-true-negative error (OTN); for every metric, lower is better. We
report \emph{means} over the $5$ replications for simulated conditions, and the
per-dataset values for real data.

\subsection{Experimental Results Based on Simulated Data}
\label{sec:results:sim}

\paragraph{Overall accuracy.}
Table~\ref{tab:sim_by_rho} reports the mean and replication standard
deviation of each metric, both overall and broken down by skill
correlation $\rho$. Across all $60$ conditions, the classical autoencoder
is the more accurate estimator, achieving an overall error of $0.133$
against the quantum autoencoder's $0.235$ (bottom row). This is expected:
DINA generates a conjunctive, bilinear item--skill structure that
coincides with the additive, tied-weight inductive bias of the CAE, and
prior work establishes that a main-effects model provably selects the
required skills under this generative model~\cite{li2021rbm}. The gap is
largest on OTP (missed loadings), where the CAE's matched structure is
most advantageous. The averages, however, hide two trends that reverse
this picture as the data becomes more realistic.

\begin{table*}[tb]
  \centering
  \setlength{\tabcolsep}{5pt}
  \begin{tabular}{@{}lcccccc@{}}
    \toprule
     & \multicolumn{3}{c}{\textbf{Classical AE (CAE)}}
     & \multicolumn{3}{c}{\textbf{Quantum Sparse (QSAE)}} \\
    \cmidrule(lr){2-4}\cmidrule(lr){5-7}
    $\rho$ & OE & OTP & OTN & OE & OTP & OTN \\
    \midrule
    $0$
      & $0.091 \pm 0.009$ & $0.148 \pm 0.023$ & $0.083 \pm 0.010$
      & $0.237 \pm \mathbf{0.007}$ & $0.459 \pm \mathbf{0.019}$ & $0.188 \pm \mathbf{0.006}$ \\
    $0.25$
      & $0.115 \pm 0.012$ & $0.136 \pm \mathbf{0.020}$ & $0.114 \pm 0.015$
      & $0.234 \pm \mathbf{0.008}$ & $0.448 \pm 0.021$ & $0.184 \pm \mathbf{0.006}$ \\
    $0.75$
      & $0.192 \pm 0.019$ & $0.220 \pm 0.036$ & $0.190 \pm 0.022$
      & $0.235 \pm \mathbf{0.007}$ & $0.463 \pm \mathbf{0.020}$ & $0.181 \pm \mathbf{0.005}$ \\
    \midrule
    Overall
      & $0.133 \pm 0.013$ & $0.168 \pm 0.026$ & $0.129 \pm 0.016$
      & $0.235 \pm \mathbf{0.007}$ & $0.457 \pm \mathbf{0.020}$ & $0.184 \pm \mathbf{0.006}$ \\
    \bottomrule
  \end{tabular}
  \caption{Simulated DINA results: mean $\pm$ replication standard
    deviation for OE, OTP, and OTN, averaged over $K$, $N$, and noise, and
    broken down by skill correlation $\rho$ (lower is better for all).
    The CAE attains lower mean error, but the QSAE is the more consistent
    estimator: within each metric, the smaller replication standard deviation is shown in
    \textbf{bold}, and the QSAE's is smaller in almost every cell.}
  \label{tab:sim_by_rho}
\end{table*}

\begin{figure}[tb]
  \centering
  \includegraphics[width=0.89\linewidth]{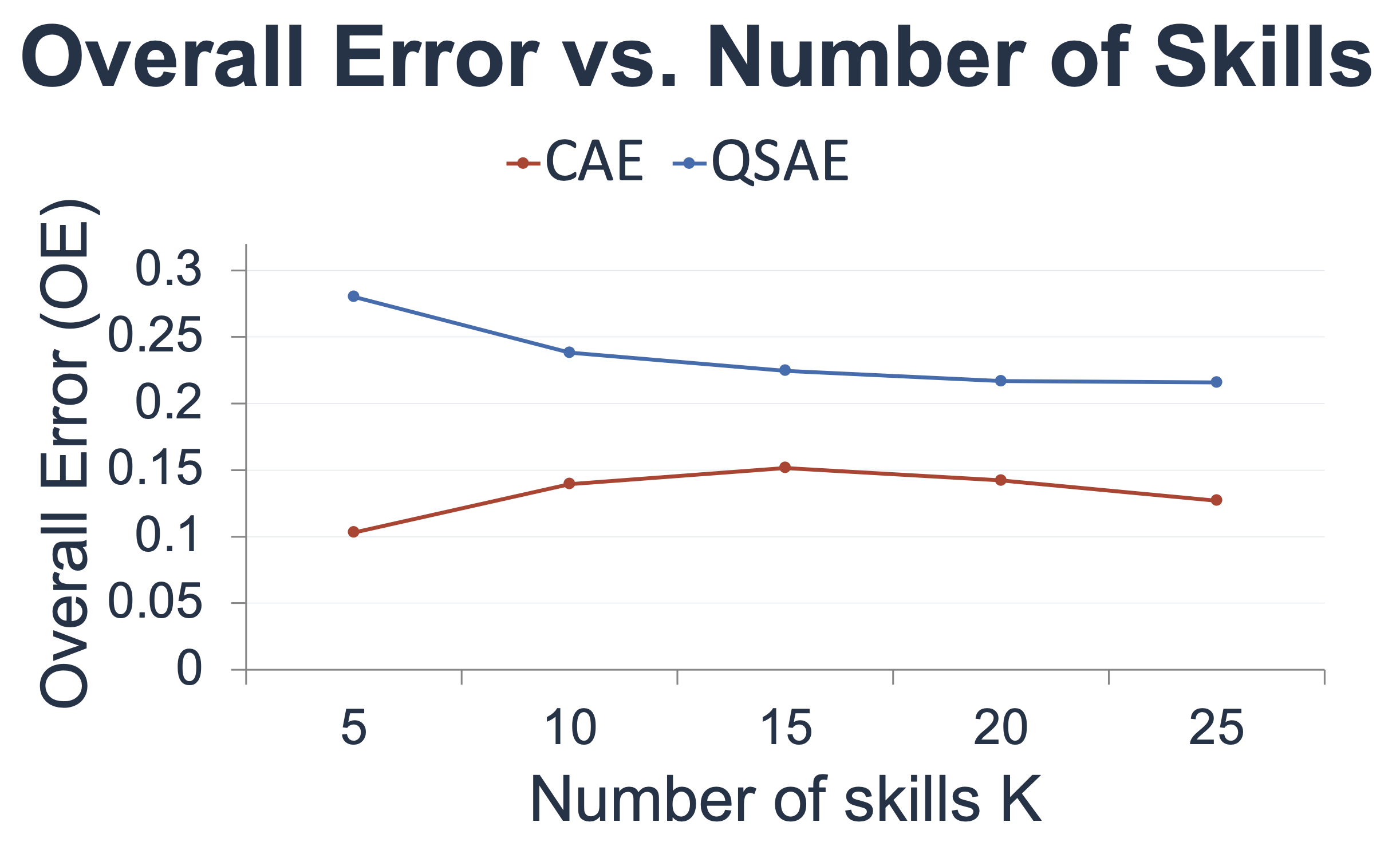}
  \caption{Overall error on simulated DINA data as a function of the number
    of skills $K$. The QSAE's (blue) error decreases as $K$ grows while the CAE's (red)
    stays flat, so the quantum--classical gap narrows steadily on larger
    problems.}
  \label{fig:oe_k}
\end{figure}

\paragraph{The quantum model scales better with more skills.}
As the number of skills $K$ grows, the quantum autoencoder's error
\emph{decreases} monotonically, from $0.280$ at $K = 5$ to $0.216$ at
$K = 25$, whereas the classical autoencoder's error stays roughly flat
between $0.10$ and $0.15$ (Figure~\ref{fig:oe_k}). The
quantum--classical gap therefore narrows steadily as the problem grows,
from $0.18$ at $K = 5$ to $0.09$ at $K = 25$. This is the regime that
matters in practice, where assessments routinely probe many skills, and it
indicates that the quantum pipeline becomes relatively more competitive
precisely as the task becomes harder.

\begin{figure}[tb]
  \centering
  \includegraphics[width=0.87\linewidth]{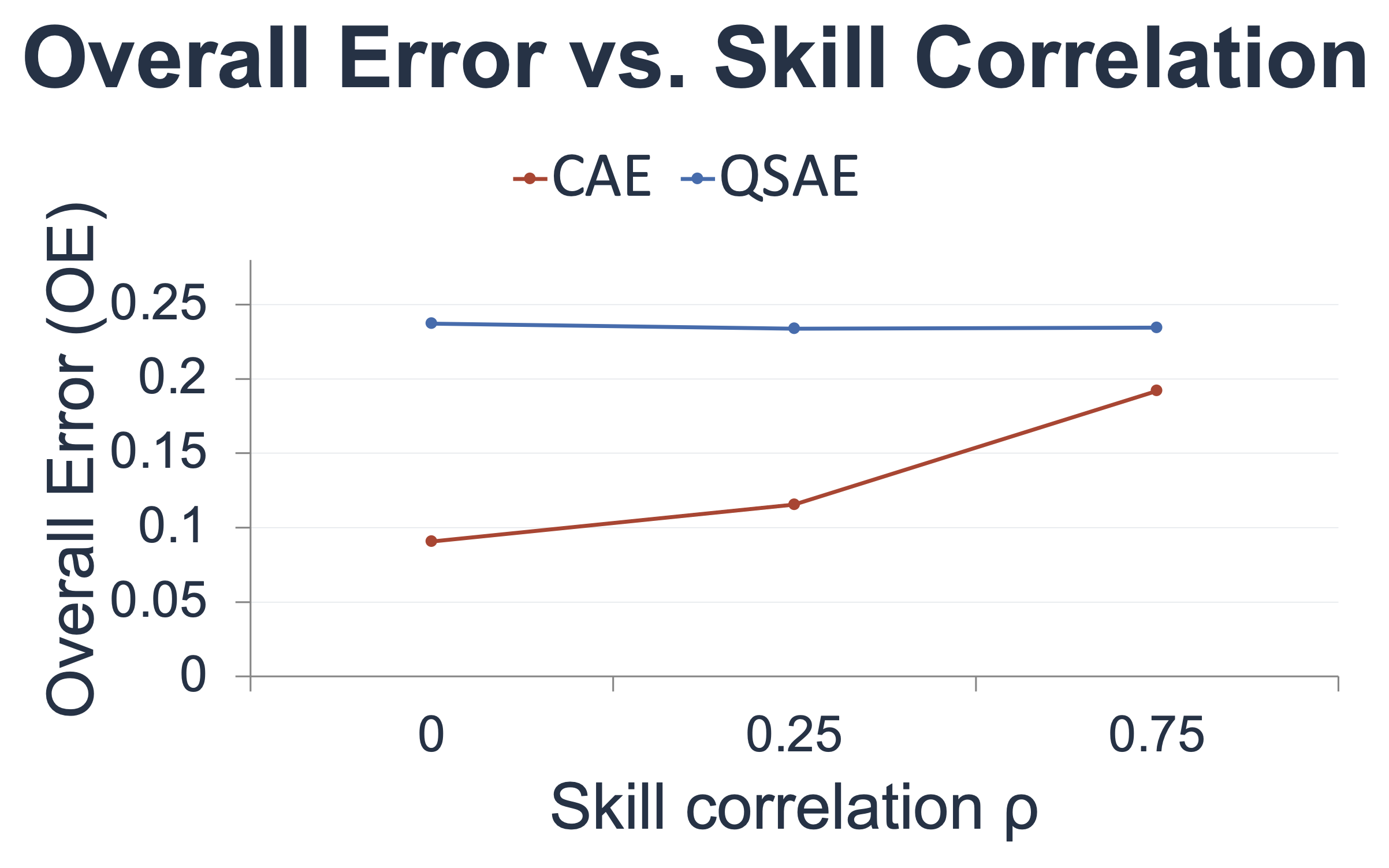}
  \caption{Overall error on simulated DINA data as a function of skill
    correlation $\rho$. The CAE (red) degrades sharply as correlation rises,
    whereas the QSAE (blue) is nearly flat, indicating that the quantum model is
    largely insensitive to correlated skills.}
  \label{fig:oe_rho}
\end{figure}

\paragraph{The quantum model is robust to skill correlation.}
The second trend is more striking. Real skills are correlated, and
correlation is the main source of difficulty in Q-matrix recovery. As
Table~\ref{tab:sim_by_rho} shows, the classical autoencoder degrades
sharply as skill correlation rises: its overall error climbs from $0.091$
in the independent case ($\rho = 0$) to $0.192$ under strong correlation
($\rho = 0.75$), more than doubling. The quantum autoencoder is, by
contrast, almost \emph{flat} across the same range, at $0.237$, $0.234$,
and $0.235$ respectively (Figure~\ref{fig:oe_rho}). Under the
strong-correlation setting that best reflects real assessments, the two
methods are far closer ($0.192$ vs.\ $0.235$) than the overall averages
suggest. The quantum representation appears largely insensitive to a
factor that systematically undermines the classical model.

\begin{figure}[tb]
  \centering
  \includegraphics[width=0.9\linewidth]{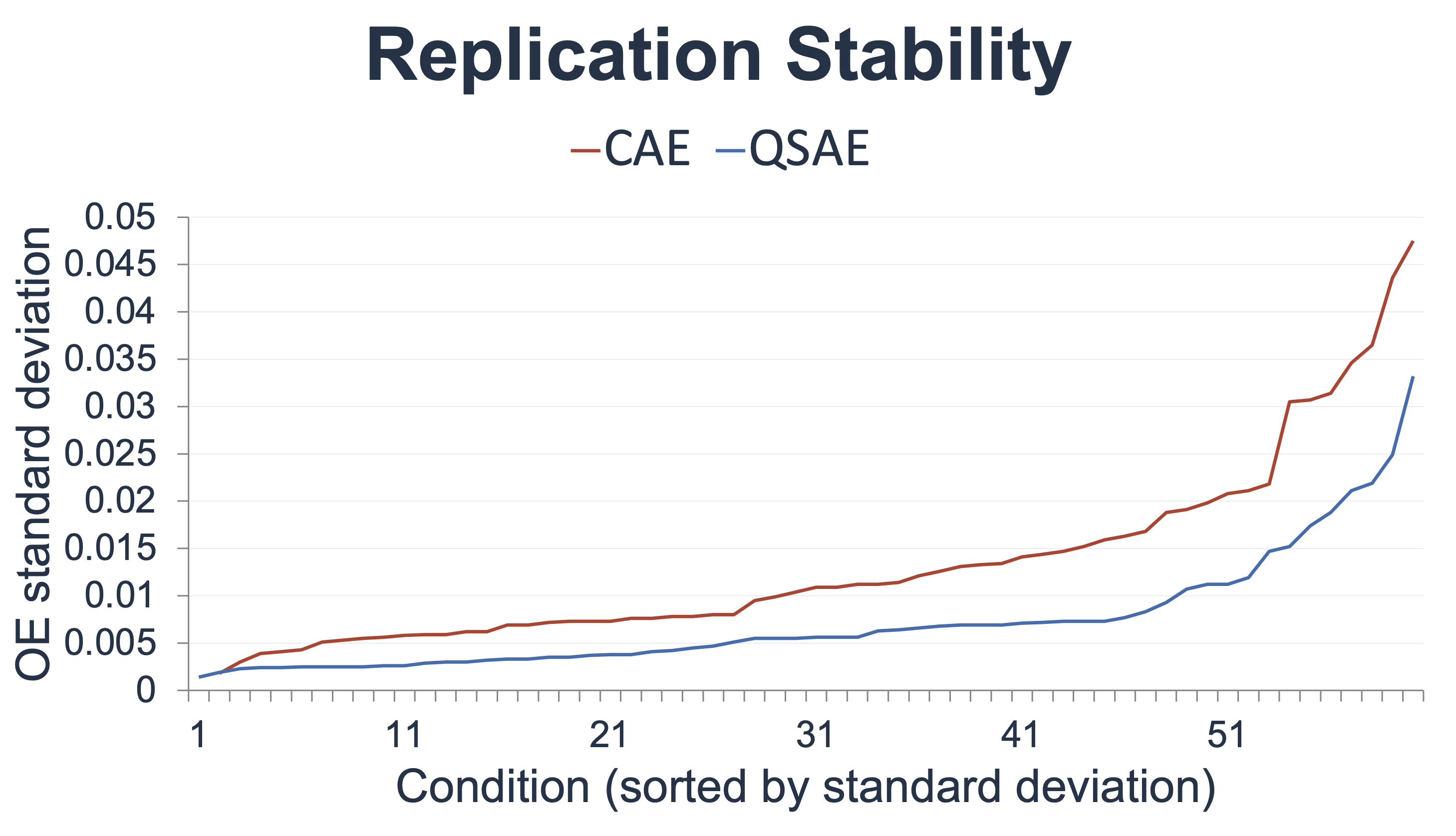}
  \caption{Per-condition replication standard deviation of OE, sorted
    within each method. The QSAE (blue) is more stable than the CAE (red)
    in $49$ of $60$ conditions.}
  \label{fig:stability}
\end{figure}

\paragraph{Stability.}
Beyond mean accuracy, the quantum autoencoder is markedly more
\emph{stable} across replications. In Table~\ref{tab:sim_by_rho}, the
smaller replication standard deviation in each metric is shown in bold,
and it falls to the QSAE in almost every cell. Aggregated over conditions,
the QSAE's OE standard deviation is lower than the CAE's in $49$ of the
$60$ conditions and roughly half as large on average ($0.007$ vs.\
$0.013$); the same pattern holds for OTN ($54/60$) and OTP ($42/60$)
(Figure~\ref{fig:stability}). Run-to-run variability directly affects how
much a single recovered Q-matrix can be trusted, so lower variance is a
practical advantage independent of mean accuracy.

\subsection{Experimental Results Based on Real Data}
\label{sec:results:real}

\begin{figure}[htbp]
  \centering
  \includegraphics[width=0.9\linewidth]{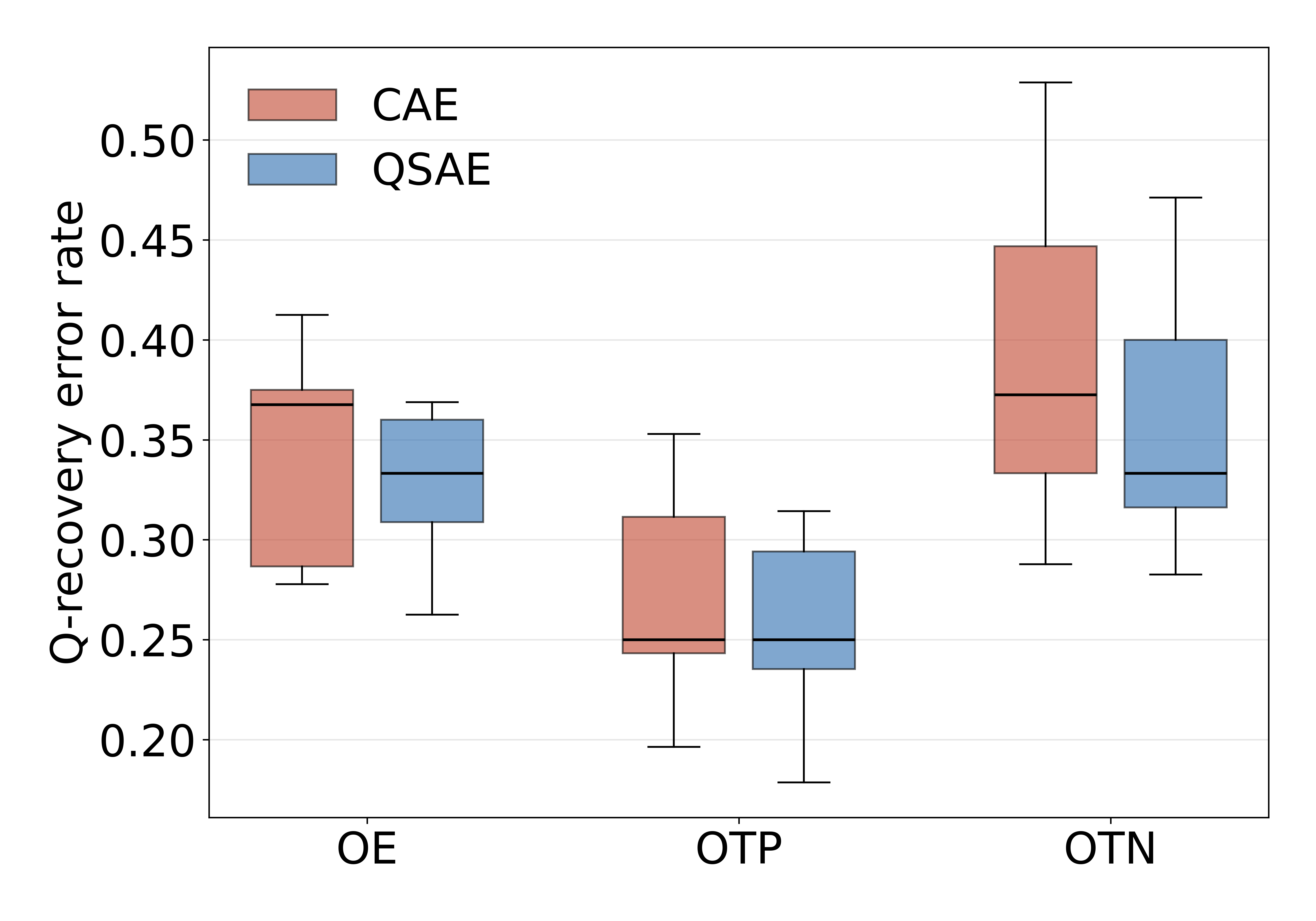}
  \caption{Distribution of the three recovery error rates (OE, OTP, OTN)
  across the $9$ real assessment datasets, comparing the classical (CAE,
  red) and quantum (QSAE, blue) autoencoders; lower is better. }
\label{fig:real_box} 
\end{figure}

On the $9$ real assessments the ordering reverses. Table~\ref{tab:real}
reports per-dataset overall error, and Figure~\ref{fig:real_box} shows the
distribution of all three metrics across datasets. Averaged across
datasets, the QSAE attains a lower mean OE than the
classical autoencoder ($0.337$ vs.\ $0.348$), and it is also lower on both
OTP ($0.272$ vs.\ $0.283$) and OTN ($0.358$ vs.\ $0.391$): the quantum
model is ahead on all three metrics. The box plot makes the shift visible: the QSAE's OE and OTN distributions sit lower and tighter than the CAE's, indicating both better central performance and fewer poor cases. At the level of individual datasets, the quantum autoencoder achieves the lower error on $6$ of the $9$ datasets, the classical autoencoder on $2$, with $1$ tie. The reversal is \emph{consistent} with the simulated findings: real assessments have unknown, non-conjunctive response processes and correlated skills, exactly the conditions under which the classical model's structural advantage disappears and the quantum model's robustness is decisive. 

\begin{table*}[tb]
  \centering
  \setlength{\tabcolsep}{8pt}
  \begin{tabular}{@{}lccccccc@{}}
    \toprule
     & \multicolumn{3}{c}{\textbf{CAE}} & & \multicolumn{3}{c}{\textbf{QSAE}} \\
    \cmidrule(lr){2-4}\cmidrule(lr){6-8}
    Dataset & OE & OTP & OTN & & OE & OTP & OTN \\
    \midrule
    dtmr\_fractions       & \textbf{0.278} & \textbf{0.257} & \textbf{0.288} & & 0.352 & 0.314 & 0.370 \\
    ecpe                  & 0.357 & 0.243 & 0.447 & & \textbf{0.333} & 0.243 & \textbf{0.404} \\
    fraction\_subtraction & 0.413 & 0.196 & 0.529 & & \textbf{0.369} & \textbf{0.179} & \textbf{0.471} \\
    hr                    & 0.371 & 0.449 & 0.341 & & \textbf{0.360} & \textbf{0.429} & \textbf{0.333} \\
    jang                  & 0.396 & 0.312 & 0.415 & & \textbf{0.309} & \textbf{0.279} & \textbf{0.316} \\
    melab                 & 0.375 & 0.235 & 0.478 & & \textbf{0.263} & 0.235 & \textbf{0.283} \\
    pgdina                & \textbf{0.287} & 0.250 & \textbf{0.311} & & 0.453 & \textbf{0.222} & 0.400 \\
    rupp\_templin\_henson & 0.286 & 0.250 & 0.333 & & 0.286 & 0.250 & 0.333 \\
    sda6                  & 0.368 & 0.353 & 0.373 & & \textbf{0.309} & \textbf{0.294} & \textbf{0.314} \\
    \midrule
    \textbf{Mean}         & 0.348 & 0.283 & 0.391 & & \textbf{0.337} & \textbf{0.272} & \textbf{0.358} \\
    \bottomrule
  \end{tabular}
  \caption{Per-dataset Q-recovery errors on the $9$ real assessments (lower is better; the better method in each metric shown in bold, ties
    unbolded). Averaged over datasets, the QSAE attains lower mean OE, OTP,
    and OTN. The QSAE outperforms the CAE per dataset on $6/9$ (OE), $5/9$ (OTP), and $6/9$ (OTN).}
  \label{tab:real}
\end{table*}

\paragraph{When Each Approach Wins.}
The CAE performs better in the regime for which its inductive bias was built: clean, independent-skill, model-matched data, where its error is lowest. The proposed QSAE overtakes the classical one in every aggregate metric and in two-thirds of individual datasets. The QSAE is thus the more robust estimator: its performance is largely invariant to the factors that degrade the classical model, and this robustness is what makes the QSAE a stronger choice in the realistic regime.

\section{Conclusion}
\label{sec:conclusion}
In this work, we introduced a novel QSAE framework
for data-driven Q-matrix estimation in educational data mining. To the best of our knowledge, this is the first application of a QML framework
to Q-matrix recovery. Our experiments demonstrate the complementary strengths of quantum and classical representation learning. Across $60$ simulated DINA conditions, the CAE generally achieves higher average Q-matrix recovery accuracy when the data closely follow the assumed generative structure. In contrast, the QSAE exhibits greater stability across replications, producing lower standard deviations in $49$ of the $60$ conditions. More importantly, on $9$ real-world assessment datasets, the QSAE outperforms the CAE on $6$ datasets. These
results suggest that the potential value of the QSAE lies not in universal
performance gains under idealized conditions, but in its robustness to
sampling variability and the more complex structure encountered in
real-world assessment data. 


\bibliography{ref}

\end{document}